\documentclass{ecai} 

\usepackage{latexsym}
\usepackage{amssymb}
\usepackage{amsmath}
\usepackage{amsthm}
\usepackage{booktabs}
\usepackage{enumitem}
\usepackage{graphicx}
\usepackage{color}

\newtheorem{theorem}{Theorem}

\newtheorem{corollary}[theorem]{Corollary}

\newtheorem{definition}{Definition}

\usepackage{amsfonts}
\usepackage{dsfont}
\usepackage{cleveref}
\usepackage{thm-restate}
\usepackage{stmaryrd}
\usepackage{tikz}

\newcommand{\ap}{\mathit{AP}}

\newcommand{\directions}{\mathbb{D}}

\newcommand{\ldot}{\mathpunct{.}}

\newcommand{\nat}{\mathbb{N}}
\newcommand{\bool}{\mathbb{B}}

\newcommand{\calT}{\mathcal{T}}

\newcommand{\calA}{\mathcal{A}}

\newcommand{\calO}{\mathcal{O}}

\newcommand{\calP}{\mathcal{P}}

\providecommand{\ltlN}{\operatorname{%
		\tikz[baseline]{
			\draw[line width=.12ex]
			(0,.6ex) circle (.8ex);
}}}{}

\providecommand{\ltlF}{\operatorname{%
		\tikz[baseline]{
			\draw[line width=.12ex,line join=round]
			(0ex,.6ex) -- (.95ex,1.55ex) -- (1.9ex,.6ex) -- (.95ex,-.35ex) -- cycle;
}}}{}

\providecommand{\ltlG}{\operatorname{%
		\tikz[baseline]{
			\draw[line width=.12ex,line join=round]
			(0ex,-.2ex) -- (0ex,1.3ex) -- (1.5ex,1.3ex) -- (1.5ex,.-.2ex) -- cycle;
}}}{}

\DeclareMathOperator{\ltlU}{\mathcal{U}}

\newcommand{\pre}{\mathit{pre}}
\newcommand{\eff}{\mathit{eff}}
\newcommand{\con}{\mathit{con}}
\newcommand{\add}{\mathit{add}}
\newcommand{\del}{\mathit{del}}
\newcommand{\apos}{\mathit{pos}}
\newcommand{\aneg}{\mathit{neg}}

\newcommand{\paths}{\mathit{Paths}}

\newif\iffullversion
\fullversiontrue

\newcommand{\ifFull}[2]{\iffullversion#1\else#2\fi}

\newcommand{\BibTeX}{B\kern-.05em{\sc i\kern-.025em b}\kern-.08em\TeX}

\begin{document}


\begin{frontmatter}

\paperid{7984} 

\title{On Conformant Planning and Model-Checking \\of $\exists^*\forall^*$ Hyperproperties}

\author{\fnms{Raven}~\snm{Beutner}}
\author{\fnms{Bernd}~\snm{Finkbeiner}}

\address{CISPA Helmholtz Center for Information Security, Germany}

\begin{abstract}
We study the connection of two problems within the planning and verification community: Conformant planning and model-checking of hyperproperties. 
Conformant planning is the task of finding a sequential plan that achieves a given objective independent of non-deterministic action effects during the plan's execution. 
Hyperproperties are system properties that relate multiple execution traces of a system and, e.g., capture information-flow and fairness policies.
In this paper, we show that model-checking of $\exists^*\forall^*$ hyperproperties is closely related to the problem of computing a conformant plan. 
Firstly, we show that we can efficiently reduce a hyperproperty model-checking instance to a conformant planning instance, and prove that our encoding is sound and complete. 
Secondly, we establish the converse direction: Every conformant planning problem is, itself, a hyperproperty model-checking task.
\end{abstract}

\end{frontmatter}


\section{Introduction}

In this paper, we identify two problems from two different research communities that seem unrelated at first glance, yet share the same computational challenge:
Conformant planning and model-checking of $\exists^*\forall^*$ hyperproperties.

\paragraph{Conformant Planning}

Conformant planning is the task of finding a plan given uncertainty about the effect of action, and without any sensing ability during the plan's execution.
That is, the same plan (i.e., sequence of actions) should achieve the goal, regardless of which non-deterministic action effects occur during the plan's execution \cite{GoldmanB96}.  

\paragraph{Hyperproperties}

Hyperproperties \cite{ClarksonS08} are system properties that relate multiple executions in a system and can thus capture properties that cannot be expressed by reasoning over individual traces.
As an example, we consider a simple information-flow property. 
Assume we model the behavior of a system as a transition system over variables $\{o, h, l\}$, and want to specify that the output ($o$) of the system does not leak information about the secret input ($h$).
We cannot specify such a property by reasoning about traces in isolation (e.g., in LTL).
Instead, we need to relate multiple executions to observe how different inputs impact the output; a hyperproperty.
HyperLTL \cite{ClarksonFKMRS14} extends LTL with quantification over executions and can thereby express hyperproperties. 
For example, we can express a simple information-flow policy -- called \emph{non-inference} (NI) \cite{McLean94} -- in HyperLTL as follows
\begin{align}\label{eq:ni}
	\forall \pi_1\ldot \exists \pi_2\ldot \ltlG (o_{\pi_1} = o_{\pi_2} \land l_{\pi_1} = l_{\pi_2}) \land \ltlG (h_{\pi_2} = \dagger), \tag*{(NI)}
\end{align}
where $\ltlG$ denotes LTL's \emph{globally} operator. 
This formula states that for any execution (trace) $\pi_1$, there exists some execution $\pi_2$ that \textbf{(1)} globally has the same low-security observations as $\pi_1$ (i.e., output $o$ and low-security input $l$ globally agree between $\pi_1$ and $\pi_2$), and \textbf{(2)} the high-security input on $\pi_2$ equals some dummy value (denoted $\dagger$).
If \ref{eq:ni} holds, an attacker thus cannot distinguish any high-security input sequence from the sequence of dummy values. 

\paragraph{Two Sides of the Same Coin}

Conformant planning and model-checking of HyperLTL are computationally expensive problems within their respective communities.
At first glance, they seem unrelated, and diverse solution concepts exist in both communities. 
Techniques employed in conformant planning often rely on an efficient heuristic search over \emph{belief states}, i.e., sets of states that contain exactly those states that are currently plausible \cite{HoffmannB06,BonetG00}. 
In contrast, techniques employed in model-checking of hyperproperties employ more direct approaches using automata \cite{BeutnerF23}, symbolic execution \cite{CorrensonNFW24}, program logics \cite{DardinierM24,CorrensonF25}, or bounded unrolling \cite{HsuSB21}.
Despite their differences, we demonstrate that the core algorithmic challenge between the two problems is shared. 
Concretely, we present efficient translations that reduce a conformant planning problem to an equivalent HyperLTL model-checking problem, and vice versa. 
Our ultimate hope is that this observation will lead to new solutions by adapting successful concepts from conformant planning to HyperLTL model-checking, and vice versa. 

\paragraph{Hyperproperty Model-Checking as Conformant Planning}

Our first contribution is a novel encoding of HyperLTL verification into conformant planning.
Our encoding is applicable to $\exists^*\forall^*$ HyperLTL formulas, i.e., formulas where an arbitrary number of existential trace quantifiers is followed by an arbitrary number of universal quantifiers. 
We can massage every formula with at most one quantifier alternation into a $\exists^*\forall^*$ formula. 
For example, \ref{eq:ni} is a $\forall\exists$ formula, but we can simply check the \emph{negated} formula, which is a $\exists\forall$ formula:
\begin{align}\label{eq:ni-neg}
	\exists \pi_1\ldot \forall \pi_2\ldot \ltlF (o_{\pi_1} \neq o_{\pi_2} \! \lor \! l_{\pi_1} \neq l_{\pi_2}) \lor \ltlF (h_{\pi_2} \neq \dagger) \tag*{(NI$_\neg$)}
\end{align}
where $\ltlF$ denotes LTL's \emph{eventually} operator. 
To check if some transition system $\calT$ satisfies \ref{eq:ni-neg}, we view model-checking as a conformant planning problem.
Each state in the planning problem maintains two system locations of $\calT$, one for path $\pi_1$ and one for $\pi_2$. 
The idea is that each plan (i.e., sequence of actions) should define a unique path for $\pi_1$, i.e., each action updates the location for $\pi_1$ along some transition of $\calT$.
At the same time, the location of $\pi_2$ is updated \emph{non-deterministically}. 
Any plan, therefore, defines a unique witness path for $\pi_1$, while the plan's executions non-deterministically explore \emph{all} possible paths for $\pi_2$. 
The planning goal is to ensure that all executions of the plan eventually reach a goal state where  $o_{\pi_1} \neq o_{\pi_2} \lor l_{\pi_1} \neq l_{\pi_2}$ or $h_{\pi_2} \neq \dagger$. 
We prove that if the resulting planning instance admits a conformant plan, $\calT$ satisfies \ref{eq:ni-neg}. 
Note how the conformant nature of the plan (i.e., the fact that the plan cannot depend on the nondeterministic outcomes) is critical to ensure that the witness trace for $\pi_1$ does not depend on $\pi_2$.
Crucially, our encoding is not only sound (if a conformant plan exists, the HyperLTL formula is satisfied) but also \emph{complete}, i.e., if $\calT$ satisfies \ref{eq:ni-neg}, there exists a conformant plan. 
This is in sharp contrast to previous planning-based encodings of hyperproperties \cite{BeutnerF24} (we discuss this in \Cref{sec:related-work}).
We show that our encoding works both with an explicit-state representation (as in \cite{BeutnerF24}), but also applies to symbolically represented planning problems (e.g., STRIPS planning).

\paragraph{Conformant Planning is a Hyperproperty}

After demonstrating that we can use conformant planning for model-checking $\exists^*\forall^*$ hyperproperties, we prove that conformant planning itself \emph{is} a hyperproperty.
Conformant planning requires a plan that achieves the goal independent of the effects of non-deterministic actions. 
The existence of a conformant plan thus corresponds to the satisfaction of the (informal) $\exists\forall$ HyperLTL formula $\exists \pi_1\ldot \forall \pi_2\ldot \big(\ltlG \mathit{sameAction}(\pi_1, \pi_2)\big) \to \ltlF \mathit{goal}(\pi_2)$ over a transition system that generates all possible paths in the planning problem. 
I.e., there exists some path $\pi_1$, such that all paths $\pi_2$ with the same sequence of actions ($\ltlG \mathit{sameAction}(\pi_1, \pi_2)$) eventually reach the goal ($\ltlF \mathit{goal}(\pi_2)$). 
While the close connection between conformant plans and quantification has been explored extensively before (e.g., in the form of SAT or QBF encodings \cite{Rintanen99,Rintanen07}; cf.~\Cref{sec:related-work}), HyperLTL can directly capture the temporal nature of plans \emph{without} bounding the length.
We implement this translation from PDDL to HyperLTL model-checking instances in a prototype.
Our results show that current HyperLTL verification tools struggle with the resulting instances, thus \textbf{(1)} creating a challenging set of benchmarks for future evaluation, and \textbf{(2)} highlighting the importance of heuristics in model-checking of $\exists^*\forall^*$ hyperproperties (an entirely unexplored research area).

\paragraph{Supplementary Materials}

Full proofs of all results can be found in the \ifFull{appendix}{full version \cite{fullVersion}}. 

\section{Related Work}\label{sec:related-work}

\paragraph{Hyperproperty Model-Checking}

Finite-state model-checking of HyperLTL is decidable \cite{ClarksonFKMRS14} but expensive: checking a formula with $k$ quantifier alternations is $k$-fold exponential \cite{Rabe16}.
Complete algorithms rely on expensive automata complementation or language inclusion checks \cite{FinkbeinerRS15,BeutnerF23}.
Approximations for this expensive problem include QBF-based bounded unrolling \cite{HsuSB21}, or a strategy-based instantiation of existential quantification \cite{CoenenFST19,BeutnerF22,BeutnerF22a}. 
In the former, the system is unrolled up to a fixed depth, so the quantification over traces reduces to a QBF formula. 
In the latter, we interpret verification of a $\forall^*\exists^*$ formula  as a game between the universal and existential quantifiers, and attempt to find a winning strategy for the existential player. 
In contrast to these abstractions, our planning-based encoding precisely captures the HyperLTL semantics (i.e., it is sound and complete), at the cost of encoding into another computationally expensive problem: conformant planning \cite{Bonet10}. 

\paragraph{Hyperproperties and Planning}
The connection between (classical) planning and formal verification has been explored in various forms \cite{GnadELH20,GiunchigliaT99,EdelkampH01,CimattiPRT03}. 
The work most closely related to ours is \cite{BeutnerF24}, showing that every HyperLTL model-checking problem can be translated soundly to a (contingent) multi-agent planning problem represented as a QDecPOMDP \cite{BrafmanSZ13}.
In \cite{BeutnerF24}'s approach, verifying a $\forall^*\exists^*$ property abstracts to a FOND-planning problem by searching for a policy that resolves existentially-quantified traces, similar to the game-based verification approaches discussed above \cite{CoenenFST19,BeutnerF22}.
This abstraction is \emph{incomplete} (cf.~\cite[Remark 1]{BeutnerF24}), i.e., a property might hold, but no witnessing contingent plan exists. 
The encoding of the present paper uses related ideas (i.e., we also simulate multiple paths in a planning problem) but identifies sensorless behavior -- i.e., \emph{conformant} instead of (fully-observable) \emph{contingent} planning -- as the key missing gadget.
Consequently, our encoding for $\exists^*\forall^*$ properties (which by negation also applies to $\forall^*\exists^*$ formulas) is sound-and-\emph{complete}.
Conformant planning is thus a drop-in replacement for model-checking $\exists^*\forall^*$ properties; contingent (FOND) planning is only an abstraction. 
Moreover, we also study the reverse direction and show that conformant planning is, itself, a $\exists\forall$ hyperproperty. 
Obtaining a similar result for the QDecPOMDP seems challenging.

\paragraph{Knowledge and Games}
The connection between (missing) knowledge of an agent and verification of hyperproperties has been explored extensively in the context of game-based verification for HyperLTL \cite{BeutnerF25a,BeutnerF25b,Winter025}.
We can view conformant planning as a special form of a two-player game, played between the planning agent and the environment, without any observations. 
In this light, we can see our planning-based encoding as a specialized verification game for $\exists^*\forall^*$ properties (compared to, e.g., the game from \cite{BeutnerF25a,BeutnerF25b}). 
A clear advantage to a planning-based approach is the fact that existing planning frameworks and tools often study symbolically represented domains (e.g., STRIPS or PDDL planning), whereas formal games under imperfect information are mostly studied in an explicit-state setting.

\paragraph{Conformant Planning and QBF}
Most conformant planning approaches employ heuristic search over \emph{belief states}, i.e., sets of world states \cite{BonetG00,HoffmannB06}, represented, e.g., as CNFs or BDDs \cite{CimattiR00}.
Another line of research related to the present paper are QBF-based approaches to conformant planning \cite{Rintanen99,Rintanen07}.
Similar to our encoding into HyperLTL, these approaches also exploit the connection between $\exists\forall$ quantification and conformant plans, either by directly using QBF or by repeated queries to a SAT solver. 
Given the temporal nature of the planning problem, a QBF encoding usually employs a bound on the length of the plan, similar to SAT-based classical planning \cite{KautzS96}.
In our framework, we can encode the temporal requirements directly using temporal logics (HyperLTL), allowing us to encode \emph{unbounded} reachability properties or even more complex temporal requirements. 
The resulting HyperLTL model-checking query can then be handled by a wide range of techniques, some of which, themself, rely on QBF-solving \cite{HsuSB21}. 

\section{Preliminaries}\label{sec:prelim}

In this section, we introduce planning problems, transition systems, and HyperLTL.
Planning problems are typically defined in a factored representation (e.g., STRIPS, PDDL) using Boolean propositions (or fluents) to represent the current state of the planning problem. 
Likewise, systems are typically symbolically defined by a set of Boolean variables (e.g., circuits or NuSmv models).
To begin, we work with an \emph{explicit-state} representation of the planning domain and system, simplifying our encoding.
In \Cref{sec:symbolic}, we then show how we can extend our encodings to symbolically represented planning problems and systems. 

\subsection{Conformant Planning}\label{sec:planning}

\begin{definition}\label{def:plan}
	A (non-deterministic) planning problem is a tuple $\calP = (S, s_0, G, \calO)$, where $S$ is a finite set of states, $s_0 \in S$ is an initial state, $G \subseteq S$ is a set of goal states, and $\calO$ is a finite set of actions. 
	Each action $a \in \calO$ has the form $a = \langle \pre_a, \eff_a\rangle$, where $\pre_a \subseteq S$ is a set of states defining the states in which the action can be applied, and $\eff_a : \pre_a \to (2^S \setminus \{\emptyset\})$ is the non-deterministic effect function mapping each state $s \in \pre_a$ to a non-empty set of potential outcomes $\eff_a(s) \subseteq S$.
\end{definition}

Note how our definition assumes that there exists a unique initial state, which is w.l.o.g., as in our setting, action effects can be non-deterministic.
The update function $\eff_a$ easily allows us to model conditional effects, which are crucial in conformant planning \cite{HoffmannB06}. 

A plan is then a sequence of actions $\langle a_1, \ldots, a_n \rangle$. 
Given a set of states $T \subseteq S$, we inductively define $\llbracket T, \langle a_1, \ldots, a_n \rangle \rrbracket \subseteq S$ as the set of states reachable after executing plan $\langle a_1, \ldots, a_n \rangle$. 
For the empty plan $\langle \rangle$, we have $\llbracket T, \langle \rangle \rrbracket := T$. 
For a non-empty plan we define 
\begin{align*}
	\llbracket T, \langle a_1, \ldots, a_n \rangle \rrbracket := \llbracket \bigcup_{s \in T}\eff_{a_1}(s) , \langle a_2, \ldots, a_n \rangle \rrbracket
\end{align*}
if $T \subseteq \pre_{a_1}$. 
That is, we consider all possible states $s \in T$ and all possible effects when applying $a_1$ ($\bigcup_{s \in T}\eff_{a_1}(s)$), and then inductively execute the remaining plan $\langle a_2, \ldots, a_n \rangle$. 
Note that $\llbracket T, \langle a_1, \ldots, a_n \rangle \rrbracket$ is undefined if action $a_1$ is not applicable in some state in $T$ \cite{HoffmannB06}.
A plan $\langle a_1, \ldots, a_n \rangle$ is a \emph{conformant plan} if $\llbracket \{s_0\}, \langle a_1, \ldots, a_n \rangle \rrbracket$ is defined and $\llbracket \{s_0\}, \langle a_1, \ldots, a_n \rangle \rrbracket \subseteq G$.
That is, executing the plan from the initial state $s_0$ always results in a goal state, independent of the non-deterministic action effects.
We assume that for every $s \in G$ and $a \in \calO$ with $s \in \pre_a$, $\eff_a(s) \subseteq G$.

\subsection{Transition Systems}

We assume that $\ap$ is a fixed set of \emph{atomic propositions}.
As the basic system model, we use finite-state transition systems (TS). 

\begin{definition}\label{def:ts-dir}
	A transition system (TS) is a tuple $\calT = (L, l_\mathit{init}, \allowbreak \directions, \kappa, \ell)$, where $L$ is a finite set of locations (following \cite{BeutnerF24}, we use ``locations'' to distinguish them from planning ``states''), $l_\mathit{init} \in L$ is an initial location, $\directions$ is a finite set of \emph{directions},  $\kappa : L \times \directions \to L$ is a transition function, and $\ell : L \to 2^\ap$ is a labeling.
\end{definition}

A path in $\calT$ is an infinite sequence $p \in L^\omega$ of locations such that \textbf{(1)} $p(0) = l_\mathit{init}$, and \textbf{(2)} for every $i \in \nat$, there exists some direction $d \in \directions$ with $p(i+1) = \kappa(p(i), d)$.
We define $\paths(\calT) \subseteq L^\omega$ as the set of all paths in $\calT$.

\subsection{HyperLTL}

HyperLTL extends LTL with explicit quantification over system executions \cite{ClarksonFKMRS14}, thus lifting it from a logic expressing trace properties to one expressing hyperproperties. 
In this paper, we focus on $\exists^*\forall^*$ formulas, i.e., formulas where any number of existential quantifiers is followed by any number of universal quantifiers. 
Such formulas are generated by the following grammar
\begin{align*}
	\psi &:= a_{\pi_i} \mid \psi \land \psi \mid \neg \psi \mid \ltlN \psi  \mid \psi \ltlU \psi \\
	\varphi &:=\exists \pi_1\ldots \exists \pi_n. \forall \pi_{n+1}\ldots \forall \pi_{n+m}\ldot \psi
\end{align*}
where $\pi_1, \ldots, \pi_{n+m}$ are so-called \emph{path variables}, and $a \in \ap$ is an atomic proposition. 
Here $\ltlN$ and $\ltlU$ denote LTL's \emph{next} and \emph{until} operator, respectively. 
Within the LTL body, we use the usual derived boolean constants and connectives $\mathit{true}, \mathit{false}, \lor, \to, \leftrightarrow$, and the temporal operators  \emph{eventually} ($\ltlF \psi := \mathit{true} \ltlU \psi$), and \emph{globally} ($\ltlG \psi := \neg \ltlF \neg \psi$).
To define the semantics, we use a path assignment $\Pi : \{\pi_1, \ldots, \pi_{n+m}\} \to \paths(\calT)$, which maps each path variable to a path. 
Given a path assignment $\Pi$, we can evaluate the LTL body $\psi$ at some position $i \in \nat$ as follows:
\begin{align*}
	\Pi, i &\models a_\pi &\text{iff } \quad&a \in \ell\big( \Pi(\pi)(i) \big)\\
	\Pi, i&\models \psi_1 \land \psi_2 \!\!\!\!\!\!\!\!&\text{iff } \quad &\Pi,i \models \psi_1 \text{ and } \Pi, i \models \psi_2\\
	\Pi, i &\models \neg \psi &\text{iff } \quad &\Pi, i \not\models \psi\\
	\Pi, i &\models \ltlN \psi &\text{iff } \quad &\Pi, i+1 \models \psi\\
	\Pi, i&\models \psi_1 \ltlU \psi_2 \!\!\! &\text{iff } \quad &\exists k \geq i\ldot \Pi, k \models \psi_2 \text{ and }  \\
	&&&\quad\quad\forall i \leq j < k\ldot \Pi, j \models \psi_1 \span \span
\end{align*}
Boolean and temporal operators are evaluated as for LTL by updating the current evaluation position $i$.
The atomic formula $a_\pi$ holds whenever $a$ holds in the current position $i$ on the path bound to $\pi$ (as given by $\calT$'s labeling $\ell$).
The quantifier prefix then quantifies over paths in the system to construct a path assignment, and evaluates this assignment on the LTL formula.
Formally, $\calT \models \exists \pi_1\ldots \exists \pi_n. \forall \pi_{n+1}\ldots \forall \pi_{n+m}\ldot \psi$, iff
\begin{align*}
	 &\exists p_1, \ldots, p_n \in \paths(\calT). \forall p_{n+1}, \ldots, p_{n+m} \in \paths(\calT)\ldot \\
	 &\quad\quad\quad\quad\quad\quad\quad\quad\quad\quad\big[\pi_1 \mapsto p_1, \ldots, \pi_{n+m} \mapsto p_{n+m}\big], 0 \models \psi.
\end{align*}
For more details on HyperLTL, we refer to \cite{Finkbeiner23}.

\section{Hyperproperty Model-Checking as Planning}\label{sec:hyper-to-conf}

In this section, we show that we can interpret the verification of an $\exists^*\forall^*$ HyperLTL formula as a conformant planning problem.
For this, assume that $\calT = (L, l_\mathit{init}, \directions, \kappa, \ell)$ is a fixed TS and $\varphi = \exists \pi_1\ldots \exists \pi_n. \forall \pi_{n+1}\ldots \forall \pi_{n+m}\ldot \psi$ is a fixed $\exists^*\forall^*$ HyperLTL formula. 
We want to check if $\calT \models \varphi$.
As sketched in the introduction, our main idea is to construct a planning problem such that each plan corresponds to concrete paths for $\pi_1, \ldots, \pi_n$.
For the plan to be successful, the concrete paths generated by that plan should be valid choices for the existentially quantified paths in $\varphi$.
That is, the paths satisfy $\psi$ (the LTL body of $\varphi$) no matter what paths we pick for $\pi_{n+1}, \ldots, \pi_{n+m}$. 
We ensure the latter by exploring \emph{all} possible choices for $\pi_{n+1}, \ldots, \pi_{n+m}$ using non-deterministic action effects. 

\paragraph{Temporal Reachability}

For now, we assume that $\psi$ -- the LTL body of $\varphi$ -- expresses a reachability property.
Note that most properties studied in practice have the form $\forall^*\exists^* \ltlG \psi$ for some propositional formula $\psi$ (see, e.g., \cite{Finkbeiner23,BeutnerF22a,CorrensonF25}), so their negation will have the form $\exists^*\forall^*. \ltlF  \neg \psi$, which is a reachability property.
We will later discuss how we can encode arbitrary $\exists^*\forall^*$ properties using more expressive planning objectives (beyond reachability).

\paragraph{DFA}
To track the reachability property expressed by $\psi$, we use a deterministic finite automaton (DFA). 
This automaton tracks whether a word satisfies the LTL body of $\varphi$, and thus operates on letters from alphabet $2^{\ap \times \{\pi_1, \ldots, \pi_{n+m}\}}$ (recall that the atoms in the LTL formula $\psi$ have the form $a_{\pi_i} \in \ap \times \{\pi_1, \ldots, \pi_{n+m}\}$).
Formally, a DFA is a tuple $\calA = (Q, q_0, \delta, F)$ where $Q$ is a finite set of states, $q_0 \in Q$ is an initial state, $\delta$ maps each pair $(q, q') \in Q \times Q$ to a Boolean formula over $\ap \times \{\pi_1, \ldots, \pi_{n+m}\}$, and $F \subseteq Q$ is a set of accepting states.
When reading a letter $\sigma \in 2^{\ap \times \{\pi_1, \ldots, \pi_{n+m}\}}$, we can transition from $q$ to $q'$ iff the evaluation defined by $\sigma$ (i.e., the assignment mapping $(a, \pi_i) \in \ap \times \{\pi_1, \ldots, \pi_{n+m}\}$ to true iff $(a, \pi_i) \in \sigma$) satisfies $\delta(q, q')$, written $\sigma \models \delta(q, q')$. 
As $\calA$ is deterministic, we can assume that for every $q \in Q$ and every $\sigma \in 2^{\ap \times \{\pi_1, \ldots, \pi_{n+m}\}}$, there exists a unique $q' \in Q$ with $\sigma \models \delta(q, q')$.
Each infinite word $u \in (2^{\ap \times \{\pi_1, \ldots, \pi_{n+m}\}})^\omega$ thus generates a unique run of $\calA$. Formally, we define $\mathit{run}_{\calA, u}\in Q^\omega$ as the unique run with $\mathit{run}_{\calA, u}(0) = q_0$, and for every $i \in \nat$, $u(i) \models \delta_\psi(\mathit{run}_{\calA, u}(i), \mathit{run}_{\calA, u}(i+1))$.
The word $u$ is accepted by $\calA$ if the unique run $\mathit{run}_{\calA, u}$ eventually reaches some state in $F$.
In the following, we assume that $\calA_\psi = (Q_\psi, q_{0, \psi}, \delta_\psi, F_\psi)$ is a DFA that accepts exactly those infinite words that satisfy $\psi$.
We can assume, w.l.o.g., that all accepting states in $F_\psi$ are sink states with a self-loop.

\paragraph{Planning Encoding}
We can now define a conformant planning problem associated with $\calT, \varphi$.

\begin{definition}\label{def:enc1}
	Define the conformant planning problem $\calP_{\calT, \varphi} := \big(S, s_0,G, \calO\big)$, where
	\begin{align*}
		S &:= \big\{\langle l_1, \ldots, l_{n+m}, q \rangle \mid  l_1, \ldots, l_{n + m} \in L, q \in Q_\psi \big\},\\
		s_0 &:= \langle l_\mathit{init}, \ldots, l_\mathit{init}, q_{0, \psi} \rangle,\\
		G &:= \big\{\langle l_1, \ldots, l_{n+m}, q \rangle \mid  l_1, \ldots, l_{n+m} \in L, q \in F_\psi\big\}, \\
		\calO &:= \{ \langle \pre_{\vec{d}}, \eff_{\vec{d}} \rangle \mid \vec{d} \in \directions^n   \}
	\end{align*} 
	and for each action $\vec{d} = (d_1, \ldots, d_n) \in \directions^n$ we define $\pre_{\vec{d}} := S$ and $\eff_{\vec{d}} : S \to S$ by
	\begin{align*}
		&\eff_{\vec{d}}\big(\langle l_1, \ldots, l_{n+m}, q \rangle\big) := \\
		&\quad\Big\{ \langle \kappa(l_1, d_1), \ldots, \kappa(l_{n+m}, d_{n+m}), q' \rangle \mid d_{n+1}, \ldots, d_{n+m} \in \directions \, \land \\
		&\quad\quad\quad \Big(\bigcup_{i=1}^{n+m} \big\{ (a, \pi_i) \mid a \in \ell(l_i)  \big\} \Big)\models \delta_\psi(q, q') \Big\}.
	\end{align*}
\end{definition}

The idea behind our definition is that the planning instance will simulate $n+m$ paths (for the path variables $\pi_1, \ldots, \pi_{n+m}$ used in $\varphi$) incrementally (guided by a plan).
For this, each planning state $\langle l_1, \ldots, l_{n+m}, q \rangle$ tracks the current location of each path ($l_1, \ldots, l_{n+m}$) and the current state of $\calA_\psi$ (thus tracking whether the simulated paths satisfy $\psi$).
We start each $\pi_i$ in the initial location $l_\mathit{init}$ and start the run of $\calA_\psi$ in the initial state $q_{0, \psi}$. 
The goal consists of all states where the automaton has reached one of $\calA_\psi$'s accepting states. 
The crux is that the actions only define the behavior of the existentially quantified paths $\pi_1, \ldots, \pi_n$, while non-determinism determines the state sequence for universally quantified paths ($\pi_{n+1}, \ldots, \pi_{n+m}$). 
Formally, each action in $\calP_{\calT, \varphi}$ is a vector $\vec{d} = (d_1, \ldots, d_n) \in \directions^n$ of $n$ directions (matching the number of existentially quantified path variables in $\varphi$).
Each action can be applied in all states (i.e., $\pre_{\vec{d}} = S$). 
When applying action $\vec{d}$ in a state $\langle l_1, \ldots, l_{n+m}, q \rangle$, locations $l_1, \ldots, l_n$ (i.e., the locations that correspond to existentially quantified paths) are updated by following the directions $d_1, \ldots, d_n$ (i.e., the $i$th location is updated to $\kappa(l_i, d_i)$). 
In contrast, the locations of universally quantified paths ($l_{n+1}, \ldots, l_{n+m}$) are updated non-deterministically, i.e., we consider all possible directions $d_{n+1}, \ldots, d_{n+m}$ in the definition of $\eff_{\vec{d}}$.
Every plan can thus precisely determine the state sequence of existentially quantified paths, while universally quantified paths explore all possible paths. 

In each step, we also update the state of $\calA_\psi$ to track whether the state sequence simulated so far satisfies the LTL body $\psi$.
For each $1 \leq i \leq n+m$, we collect all APs that hold in the current location and index them with $\pi_i$, thus obtaining a letter $\bigcup_{i=1}^{n+m} \big\{ (a, \pi_i) \mid a \in \ell(l_i)  \big\}$ in $2^{\ap \times \{\pi_1, \ldots, \pi_{n+m}\}}$.
We then transition to the unique state $q' \in Q_\psi$, with $\bigcup_{i=1}^{n+m} \big\{ (a, \pi_i) \mid a \in \ell(l_i)  \big\} \models \delta_\psi(q, q')$.

Note that the size of $\calP_{\calT, \varphi}$ is polynomial in the size of $\calT$ and exponential in $n+m$ (as usual for self-compositions \cite{BartheDR04}).
Concretely, $\calP_{\calT, \varphi}$ has $\calO(|S|^{n+m})$ many states. 

\paragraph{Soundness and Completeness}

We can show that our encoding is sound, i.e., the existence of a conformant plan implies that the hyperproperty is satisfied. 
Moreover, our key contribution is the observation that \emph{conformance} (i.e., sensorless behavior) is the key technical gadget that allows us to precisely express the semantics of HyperLTL, leading to completeness:

\begin{restatable}[Soundness and Completeness]{theorem}{thsound}\label{theo:th1}
	There exists a conformant plan for $\calP_{\calT, \varphi}$ if and only if $\calT \models \varphi$. 
\end{restatable}
\begin{proof}[Proof Sketch]
	For the first direction, assume that $\langle a_1, \ldots, a_N \rangle$ is a conformant plan for $\calP_{\calT, \varphi}$.
	We obtain $n$ finite prefixes of length $N$ by simulating the directions used in each action (recall that each action $a_i$ is an $n$-tuple of directions). 
	By extending these finite prefixes into infinite paths, we obtain concrete witness paths $p_1, \ldots, p_n$ for $\pi_1, \ldots, \pi_n$. 
	Indeed, no matter what paths $p_{n+1}, \ldots, p_{n+m}$ we consider for the $m$ universally quantified paths in $\varphi$, there exists \emph{some} execution of plan $\langle a_1, \ldots, a_N \rangle$ that traverses the prefixes of $p_{n+1}, \ldots, p_{n+m}$.
	As the plan is conformant, the paths $p_1, \ldots, p_{n+m}$ together must thus visit an accepting state of $\calA_\psi$, and thus satisfy $\psi$; $p_1, \ldots, p_n$ are witnesses for $\pi_1, \ldots, \pi_n$, so $\calT \models \varphi$.

	For the second direction, assume $\calT \models \varphi$.
	As $\calT \models \varphi$, there exist witness paths $p_1, \ldots, p_n \in \paths(\calT)$ for the existentially quantified paths $\pi_1, \ldots, \pi_n$. 
	We choose some plan that -- within $\calP_{\calT, \varphi}$ -- generates exactly the paths $p_1, \ldots, p_n$, and claim that it is conformant. 
	Indeed, every execution of this plan traverses exactly paths $p_1, \ldots, p_n$ in the first $n$ system copies, and traverses some paths $p_{n+1}, \ldots, p_{n+m} \in \paths(\calT)$ in the remaining $m$ system copies. 
	As $p_1, \ldots, p_n$ are witness traces for $\varphi$, all such combinations satisfy $\psi$ and thus eventually reach an accepting state in $\calA_\psi$. 
	Every execution of the plan thus eventually visits a goal state (after a \emph{bounded} number of steps since $\calT$ is finite-state); the plan is conformant.

	A full proof can be found in \ifFull{Appendix \ref{app:dir1}}{the full version \cite{fullVersion}}.
\end{proof}

\paragraph{Beyond Reachabilty}

So far, our encoding is limited to formulas where the LTL body expresses a reachability property, as reachability (of the goal) is the standard goal description used in (conformant) planning. 
The idea underlying our encoding can also be extended to handle arbitrary temporal requirements by using more expressive goal conditions.
Our encoding could thus be easily extended to full HyperLTL, i.e., for every $\exists^*\forall^*$ HyperLTL formula, we can construct a planning problem (\emph{with LTL-defined planning objective}) that admits a conformant plan iff the HyperLTL formula is satisfied. 
There exist many approaches that study non-deterministic planning under temporal objectives (e.g., LTL) \cite{CamachoTMBM17,CamachoM19,CalvaneseGV02,PatriziLGG11}, so far mostly in a fully-observable setting.

\section{Conformant Planning as a Hyperproperty}\label{sec:conf-to-hyper}

In the previous section, we showed that we can solve $\exists^*\forall^*$ HyperLTL model-checking by viewing it as a conformant planning problem. 
In this section, we show the reverse: conformant planning is a $\exists^*\forall^*$ hyperproperty.
We, again, first work with an explicit state representation and lift our encoding to symbolic systems in \Cref{sec:symbolic}.
Let $\calP = (S, s_0, G, \calO)$ be a fixed planning problem. 

We will first construct a TS over atomic propositions $\ap := \{\mathit{act}_a \mid a \in \calO\} \cup \{\mathit{goal}\}$, whose paths precisely correspond to all possible plan executions in $\calP$. 
The atomic propositions then allow us to \textbf{(1)} access the last action played (i.e., $\mathit{act}_a$ should hold iff action $a$ was the action used in the previous step), and \textbf{(2)} determine if the current state is a goal state (via AP $\mathit{goal}$). 
To record the last action, each location will be of the form $(s, a)$, where $s \in S$ is the current planning state, and $a \in \calO$ is the action that was last played. 
Moreover, we add locations of the form $(\lightning, a)$ (where $a \in \calO$), to indicate that action $a$ was last played but was not applicable.
As we do not need to uniquely identify transitions, we omit directions and directly view the transition function as a function $\kappa : L \to 2^L \setminus \{\emptyset\}$.
Formally, we define the explicit-state TS $\calT_\calP$ as follows:

\begin{definition}\label{def:enc2}
	Define the TS $\calT_\calP := (L, l_\mathit{init}, \kappa, \ell)$, where
	\begin{align*}
		L &:= \big\{ (s, a) \mid s \in S, a \in \calO \big\} \cup \big\{(\lightning, a) \mid a \in \calO\big\},
	\end{align*}
	and $l_\mathit{init} := (s_0, a_0)$, where $a_0 \in \calO$ is an arbitrary action.
	For the transition function we define
	\begin{align*}
		\kappa \big( (s, a)  \big) &:= \Big\{ (s', a') \mid a' \in \calO \land s \in \pre_{a'} \land s' \in \eff_{a'}(s)  \Big\} \, \cup \\
		&\quad\quad\quad\quad\Big\{ (\lightning, a') \mid a' \in \calO \land s \not\in \pre_{a'} \Big\}\\
		\kappa \big( (\lightning, a) \big) &:= \big\{ (\lightning, a') \mid a' \in \calO \big\}.
	\end{align*}
	The labeling function is defined by:
		\begin{align*}
				\ell \big( s, a \big) &:= \begin{cases}\begin{aligned}
								&\big\{\mathit{act}_a, \mathit{goal}\big\}  \quad &&\text{if } s \in G\\
								&\big\{\mathit{act}_a\big\} \quad &&\text{otherwise}
							\end{aligned}
					\end{cases}\\
				\ell \big( \lightning, a\big) &:= \big\{\mathit{act}_a \big\}.
			\end{align*}
\end{definition}

It is easy to see that every sequence of state-action pairs $(s_0, a_0)(s_1, a_1)\cdots (s_n, a_n) \in (S \times \calO)^*$ is the prefix of some path in $\paths(\calT_\calP)$ iff $s_0, s_1, \ldots, s_n$ is a sequence of states in $\calP$ under plan $\langle a_1, \ldots, a_n \rangle$.
The TS $\calT_\calP$ thus generates all possible plan executions in $\calP$, and records the last actions in each location.
Note that once we reach an error location of the form $(\lightning, a)$ (by playing a non-applicable action), we can never reach a location where AP $\mathit{goal}$ holds.

\begin{definition}
	Define the HyperLTL formula $\varphi_\calP$ by
	\begin{align*}
		\varphi_\calP := \exists \pi_1\ldot \forall \pi_2\ldot \ltlF \mathit{goal}_{\pi_2} \lor  \ltlF \Big(\bigvee_{a \in \calO} (\mathit{act}_a)_{\pi_1} \not\leftrightarrow (\mathit{act}_a)_{\pi_2} \Big).
	\end{align*}
\end{definition}

That is, we require some plan (modeled as a sequence $\pi_1$ of state-action pairs) such that all paths $\pi_2$ with the same action sequence (i.e., all paths that follow the same plan as encoded in path $\pi_1$) eventually reach the goal. 
Phrased differently, any path $\pi_2$ must either reach the goal or eventually use a different action than used on $\pi_1$.
Note that $\varphi_\calP$ also implies that $\pi_1$ eventually reaches the goal; by instantiating $\pi_2$ with the same path used for $\pi_1$, the second disjunct will never be satisfied.

\begin{restatable}[Soundness and Completeness]{theorem}{second}\label{theo:th3}
	There exists a conformant plan for $\calP$ if and only if $\calT_\calP \models \varphi_\calP$. 
\end{restatable}
\begin{proof}[Proof Sketch]
	For the first direction, assume that $\calP$ admits a conformant plan $\langle a_1, \ldots, a_N \rangle$.
	To show $\calT_\calP \models \varphi_\calP$, we need to provide a witness trace for $\pi_1$. 
	We create this witness path $p_1$ by choosing any path in $\calT_\calP$ where the first $N$ actions (which are recorded in each location of $\calT_\calP$) are exactly $a_1, \ldots, a_N$. 
	To show that $p_1$ is a witness path for $\pi_1$ consider any possible path $p_2 \in \paths(\calT)$ for the universally quantified $\pi_2$ in $\varphi_\calP$. 
	There are two options: Either the first $N$ actions in $p_2$ are exactly $a_1, \ldots, a_N$, or at some position, the action differs. 
	In the latter case, the second disjunct in the body of $\varphi_\calP$ is satisfied.
	In the former case, the action sequence is exactly $\langle a_1, \ldots, a_N \rangle$, so, by construction of $\calT_\calP$, the state sequence is some execution in $\calP$ under plan $\langle a_1, \ldots, a_N \rangle$. As $\langle a_1, \ldots, a_N \rangle$ is a conformant plan, this already implies that the state sequence visits a goal state, so the first disjunct in the body of $\varphi_\calP$ is satisfied.
	
	For the reverse direction, assume that $\calT_\calP \models \varphi_\calP$, and let $p_1 \in \paths(\calT_\calP)$ be a witness path for $\pi_1$. 
	We can extract a conformant plan by projecting on the actions recorded in $p_1$. 
	To show that this plan is conformant, consider an arbitrary execution under that plan. 
	If paired with the sequence of actions, we obtain a path $p_2 \in \paths(\calT_\calP)$ which we can use for the universally quantified $\pi_2$; As $\calT_\calP \models \varphi_\calP$, and $p_1$ is a witness for $\pi_1$, $[\pi_1 \mapsto p_1, \pi_2 \mapsto p_2]$ satisfies the body of $\varphi_\calP$.
	As the action sequence of $p_1$ and $p_2$ is the same, the second disjunct in $\varphi_\calP$'s is never satisfied, so the first disjunct must hold. 
	This already implies that $p_2$ visits a goal state, and, as $\calT$ is finite-state, the length until a goal state is visited is bounded across all executions. 
	As this holds for any execution of the plan, the plan is conformant. 
	
	A full proof can be found in \ifFull{Appendix \ref{app:dir2}}{the full version \cite{fullVersion}}.
\end{proof}

\section{STRIPS Planning and Symbolic Systems}\label{sec:symbolic}

In the previous sections, we worked with an explicit-state representation of the planning domain and the transition system. 
In practice, planning problems are typically represented symbolically using (Boolean) variables (also called propositions or fluents) to describe the current state (e.g., STRIPS or PDDL). 
Converting such a planning problem to the explicit-state description used in \Cref{def:plan}, results in an exponential blowup, making it infeasible in practice. 
Likewise, many systems used in HyperLTL verification are described symbolically (e.g., circuits or NuSMV models).
In this section, we show that the ideas of the above encodings seamlessly apply to such symbolically-represented systems and planning problems, i.e., we can directly encode HyperLTL model-checking on symbolic systems as (conformant) STRIPS-style planning, and vice versa; \emph{without} first obtaining an explicit-state representation.

\subsection{STRIPS Conformant Planning}

We consider a STRIPS-like non-deterministic planning domain:

\begin{definition}\label{def:sym-plan}
	A (non-deterministic) STRIPS planning problem is a tuple $\calP = (P, I, G, \calO)$, where $P$ is a finite set of Boolean propositions (also called fluents), $I \subseteq P$ is a set of propositions (defining the initial state), $G \subseteq P$ is a set of propositions defining goal states, and $\calO$ is a finite set of actions. Each action $a \in \calO$ has the form $a = \langle \pre_a, \eff_a\rangle$, where $\pre_a \subseteq P$ defines all propositions which must hold in order for the action to be applicable, and $\eff_a = \{e_1, \ldots, e_k\}$ is a finite set of conditional effects with $e_i = \langle \con, \add, \del \rangle$, where $\con$ is a Boolean formula over $P$ (the condition), $\add \subseteq P$ is the add list, and $\del \subseteq P$ is the delete list. 
\end{definition}

A state in $\calP$ is a set $s \subseteq P$ defining which propositions are set to true. 
An action $a = \langle \pre_a, \eff_a\rangle$ can be applied in $s$ if $\pre_a \subseteq s$. 
When applying action $a$ in state $s$, we non-deterministically execute one of the (applicable) conditional effects in $\eff_a$.
Here, a conditional effect $\langle \con, \add, \del \rangle \in \eff_a$ is applicable if $s \models \con$, i.e., the Boolean formula $\con$ is satisfied by the assignment to propositions $P$ defined by $s$.
Given a state $s$ and action $a$, we define $\mathit{apply}_a(s) \subseteq 2^P$ as the effect of applying $a$ in $s$:
\begin{align*}
 \Big\{ (s  \setminus \del) \cup  \add \mid \langle \con, \add, \del \rangle \in \eff_a \land s \models \con  \Big\}.
\end{align*}
As in \Cref{sec:planning}, we extend this to plans. 
A plan is a sequence $\langle a_1, \ldots, a_n \rangle$ of actions. 
Given a set of states $T \subseteq 2^P$, we inductively define $\llbracket T, \langle a_1, \ldots, a_n \rangle \rrbracket \subseteq 2^P$ as the set of states reachable after executing plan $\langle a_1, \ldots, a_n \rangle$. 
For the empty plan $\langle \rangle$, we define $\llbracket T, \langle \rangle \rrbracket := T$. 
For a non-empty plan we define 
\begin{align*}
	\llbracket T, \langle a_1, \ldots, a_n \rangle \rrbracket := \llbracket \bigcup_{s \in T} \mathit{apply}_{a_1}(s), \langle a_2, \ldots, a_n \rangle \rrbracket
\end{align*}
if for every $s \in T$, we have $\pre_{a_1} \subseteq s$. 
As before, a plan $\langle a_1, \ldots, a_n \rangle$ is a \emph{conformant plan} if $\llbracket \{I\}, \langle a_1, \ldots, a_n \rangle \rrbracket$ is defined and for every $s \in \llbracket \{I\}, \langle a_1, \ldots, a_n \rangle \rrbracket$ we have $s \cap G \neq \emptyset$.
That is, all executions of the plan, starting in the initial state $I$, result in goal states, i.e., states where at least one goal proposition holds. 
Our definition differs slightly from the one used by Hoffmann and Brafman \cite{HoffmannB06} in that we non-deterministically pick one of the conditional effects, allowing us to model non-deterministic action effects. 
In contrast, Hoffmann and Brafman \cite{HoffmannB06} consider deterministic actions but non-deterministic initial state(s). In our definition, we can always model deterministic effects by ensuring that at most one conditional effect is applicable.

\subsection{Symbolic Transition Systems}

Similarly, we can consider a symbolic description of transition systems. 
Here, we assume that the states are represented via Boolean variables, and each direction is assigned a list of guarded commands, which indirectly bounds the branching degree of the system to $|\directions|$.

\begin{definition}\label{def:sym-ts}
	A symbolic transition system (STS) is a tuple $\calT = (X, v_\mathit{init}, \directions)$, where $X$ is a finite set of Boolean variables, $v_\mathit{init} \subseteq X$ is an initial assignment to $X$, and $\directions$ is a finite set of \emph{directions}.
	Each direction $d \in \directions$ has the form $d = (g, \apos, \aneg)$ where $g$ is a Boolean formula over $X$ (called the guard), $\apos \subseteq X$ is a set of variables which will be set to true, and $\aneg \subseteq X$ is a set of variables which will be set to false.
\end{definition}

A state is a variable evaluation $v \subseteq X$. 
A state $v'$ is a successor of $v$, written $v \to_\calT v'$, if there exists a $d = (g, \apos, \aneg) \in \directions$ such that $v \models g$ and $v' = (v \setminus \aneg) \cup \apos$.
That is, the guard to the direction is satisfied (interpreting a state $v \subseteq X$ as the obvious assignment $X \to \bool$), all positive variables in $\apos$ are set to true, and all negative variables in $\aneg$ are set to false (similar to the add and delete list in STRIPS planning).

As before, a path in $\calT$ is an infinite sequence $p \in (2^X)^\omega$ such that \textbf{(1)} $p(0) = v_\mathit{init}$, and \textbf{(2)} for every $i \in \nat$, $p(i) \to_\calT p(i+1)$.
We define $\paths(\calT) \subseteq (2^X)^\omega$ as the set of all paths in $\calT$.
Note that our definition omits atomic propositions (APs) as we can directly view the Boolean state variables as APs.
That is, we assume that the atomic formulas within the LTL body are of the form $x_{\pi_i}$ where $x \in X$ and $\pi_i \in \{\pi_1, \ldots, \pi_{n+m}\}$.
Given a HyperLTL formula $\varphi$, we define $\calT \models \varphi$ as expected. 

\subsection{Hyperproperty Model-Checking as Planning}

We can now extend our encoding from \Cref{sec:hyper-to-conf} to symbolic systems and symbolic planning domains.
Ultimately, given an STS $\calT$ and HyperLTL formula $\varphi$, we want to construct a STRIPS planning problem (cf.~\Cref{def:sym-plan}) that admits a conformant plan iff $\calT \models \varphi$ (\emph{without} first translating the STS to an explicit-state TS).

In the following, assume that  $\calT = (X, v_\mathit{init}, \directions)$ is the fixed STS and $\varphi = \exists \pi_1\ldots \exists \pi_n. \forall \pi_{n+1}\ldots \forall \pi_{n+m}\ldot \psi$ is a fixed $\exists^*\forall^*$ HyperLTL formula. 
As before, we assume that $\psi$ -- the LTL body of $\varphi$ -- expresses a reachability property.
Let $\calA_\psi = (Q_\psi, q_{0, \psi}, \delta_\psi, F_\psi)$ be a DFA over letters from $2^{X \times \{\pi_1, \ldots, \pi_{n+m}\}}$ (recall that we use the Boolean variables in $\calT$ as APs) that accepts exactly those infinite words that satisfy $\psi$.

\paragraph{Indexed Variables}
In our planning encoding, we maintain the current location of $n+m$ system copies. 
When using an explicit-state representation as in \Cref{def:enc1}, we could simply use an $(n+m)$-tuple of system locations. 
In our symbolic setting, the current location of a system is defined by a variable evaluation over $X$, so to track the $n+m$ system copies, we use an indexed set of variables.
Formally, we will represent $n+m$ system locations within each planning state by using propositions from $X \times \{\pi_1, \ldots, \pi_{n+m}\}$.
Variables $\{(x, \pi_i) \mid x \in X\}$ then define the current location of the $i$th system copy.
Given a set of variables $A \subseteq X$, and $\pi_i \in \{\pi_1, \ldots, \pi_{n+m}\}$, we define $A_{\circ \pi_i} := \{ (x, \pi_i) \mid x \in A\}$ as the indexed set of variables.
Likewise, given a Boolean formula $g$ over variables from $X$, define $g_{\circ \pi_i}$ as the formula over $X \times \{\pi_1, \ldots, \pi_{n+m}\}$ where each variable $x$ is replaced by $(x, \pi_i)$. 

\begin{definition}
	Define the STRIPS planning problem $\calP_{\calT, \varphi}$ as $\calP_{\calT, \varphi} := \big(P, I, G, \calO\big)$, where
	\begin{align*}
		P &:= (X \times \{\pi_1, \ldots, \pi_{n+m}\}) \cup Q_\psi\\
		I & := \big\{ (x, \pi_i) \mid x \in v_\mathit{init}, 1 \leq i \leq n+m  \big\} \cup \{q_{0, \psi}\}\\
		G &:= F_\psi\\
		\calO &:= \big\{ \langle \pre_{\vec{d}}, \eff_{\vec{d}} \rangle \mid \vec{d} \in \directions^n   \big\}
	\end{align*}
	and for each action $\vec{d} \in \directions^n$ we define $\pre_{\vec{d}} := \emptyset$ and the conditional effects $\eff_{\vec{d}} $ are defined by 
	\begin{align*}
		\eff_{\vec{d}} := \Big\{   e_{\vec{d} \uplus \vec{d}', q, q' } \mid \vec{d}' \in \directions^m, q, q' \in Q_\psi \Big\}.
	\end{align*}
	For each direction vector $\vec{d} \uplus \vec{d}' = (d_1, \ldots, d_{n+m}) \in \directions^{n+m}$, where $d_i = (g_i, \apos_i, \aneg_i)$, we define the conditional effect $e_{\vec{d} \uplus \vec{d}', q, q' }$ as $e_{\vec{d} \uplus \vec{d}', q, q' } := \langle \con, \add, \del\rangle$, where 
	\begin{align*}
		\con &:= q \land \delta_\pi(q, q') \land \bigwedge_{i=1}^{n+m} (g_i)_{\circ \pi_i}\\
		\add &:= \{q'\} \cup \bigcup_{i=1}^{n+m} (\apos_i)_{\circ \pi_i}\\
		\del &:= \{q\} \cup \bigcup_{i=1}^{n+m} (\aneg_i)_{\circ \pi_i}.
	\end{align*}
\end{definition}

The idea behind this encoding is similar to the explicit-state encoding in \Cref{def:enc1}: The action chosen in each step determines how the $n$ existentially quantified systems are updated, while the $m$ universally quantified systems are updated non-deterministically.
Each location of $\calT$ is defined by the Boolean variables in $X$, so our planning problem uses propositions of the form $(x, \pi_i)$ defining the value of variable $x$ in the system copy for $\pi_i$. 
Additionally, we track the current state of $\calA_\psi$, by adding each state in $Q_\psi$ as a proposition.
Initially, a proposition $(x, \pi_i)$ is set iff variable $x$ is set to true in $\calT$'s initial state.
The goal then consists of all states where some proposition in $F_\psi$ is set, i.e., where $\calA_\psi$ is in some accepting state.
Similar to the encoding in \Cref{def:enc1}, the actions consist of $n$-tuples of directions. The idea is that each action $\vec{d}  = (d_1, \ldots, d_n )\in \directions^n$ defines how each of the $n$ existentially quantified systems is updated, while the $m$ universally quantified systems are updated non-deterministically.
Once action $\vec{d}$ is fixed, the conditional effects in $\eff_{\vec{d}}$ are executed non-deterministically (if applicable).
Here, each conditional effect in $\eff_{\vec{d}}$ has the form $e_{\vec{d} \uplus \vec{d}', q, q' }$ where $\vec{d}' \in \directions^m$ gives the direction for each of the $m$ universally quantified systems (the conditional effects thus consider all possible updates to universally quantified systems). 
At the same time, we consider all possible combinations $q, q'$ to update the state of $\calA_\psi$. 
Each effect thus determines directions for all universal copies and determines how $\calA_\psi$ is updated. 
Of the conditional effects in $\eff_{\vec{d}}$, only a few are actually applicable: 
Namely, those where the guard of all directions is satisfied and $\calA_\psi$ actually moves from $q$ to $q'$.
The condition of each conditional effect ensures this: Effect $e_{\vec{d} \uplus \vec{d}', q, q' }$ is only applicable if \textbf{(1)} $q$ holds (i.e., we are currently in state $q$), \textbf{(2)} $\delta_\psi(q, q')$ (i.e., $\calA_\psi$ moves from state $q$ to $q'$ in the current system states; Recall that $\delta_\psi(q, q')$ is a Boolean formula over $X \times \{\pi_1, \ldots, \pi_{n+m}\}$ and all these variables are propositions in the planning problem), and \textbf{(3)} all the guards of all directions are satisfied. 
To express the latter condition, we assume that $\vec{d} \uplus \vec{d}' = (d_1, \ldots, d_{n+m})$ is the direction vector for all system copies (obtained by merging the $n$ directions fixed by the action and the $m$ directions fixed in the effect), and $g_i$ is the guard of the $i$th direction (recall that $g_i$ is a Boolean formula over $X$). 
In our encoding, we track the current state of the $i$th system copy via the indexed variables $\{(x, \pi_i) \mid x \in X\}$, so, within guard $g_i$, we replace each variable $x$ with $(x, \pi_i)$ (i.e., $(g_i)_{\circ \pi_i}$).
When applying the conditional effect $e_{\vec{d} \uplus \vec{d}', q, q' }$, we move $\calA_\psi$ from $q$ to $q'$, i.e., we add the proposition $q'$ and delete the proposition $q$. 
At the same time, we update each system copy based on the chosen direction, i.e., the $i$th copy is updated based on direction $d_i$.
Formally, this amounts to adding all variables set to true by $d_i$ (i.e., adding $(\apos_i)_{\circ \pi_i}$) and removing all variables set to false by $d_i$ (i.e., deleting $(\aneg_i)_{\circ \pi_i}$).

The resulting (STRIPS-like) planning problem models the same behavior as the explicit-state construction from \Cref{def:enc1}.
As a result, we get the following direct corollary of \Cref{theo:th1}:

\begin{corollary}[Soundness and Completeness]
	There exists a conformant plan for $\calP_{\calT, \varphi}$ if and only if $\calT \models \varphi$.
\end{corollary}

Note that we can construct $\calP_{\calT, \varphi}$ in polynomial time in the size $\calT$.
Concretely, the number of fluents is $\calO((n+m) \cdot |X|)$, the number of actions is $\calO(|\directions|^n)$, and the overall number of conditional effects is $\calO(|\directions|^{n+m})$ (as usual for a self-compositions \cite{BartheDR04}).

\subsection{Conformant Planning as a Hyperproperty}\label{sec:sym2}

We now sketch how we can extend the encoding from \Cref{sec:conf-to-hyper} to  STRIPS planning problems and symbolically transition systems. 
Given a STRIPS planning problem $\calP = (P, I, G, \calO)$ (cf.~\Cref{def:sym-plan}), we can construct an STS $\calT_\calP$ (cf.~\Cref{def:sym-ts}) that has the same behavior as the explicit-state construction in \Cref{def:enc2}. 
To accomplish this, we track each proposition in $P$ as a variable in $\calT_\calP$.
Moreover, we add additional variables $\{\mathit{act}_a \mid a \in \calO\}$ (used to record the last action that was played), $\mathit{goal}$ (set whenever we reach a goal proposition from $G$), and $\lightning$ (set whenever we have previously played a non-applicable action, similar to the $(\lightning, a)$ states in \Cref{def:enc2}).
The initial state of $\calT_\calP$ sets exactly those propositions contained in $I$ (i.e., $\calP$'s initial state).
For each action $a = \langle \pre_a, \eff_a \rangle \in \calO$ with $k$ conditional effects $\eff_a = \{e_1, \ldots, e_k\}$, we add $k$ directions $d_{(a, e_1)}, \ldots, d_{(a, e_k)}$ and a special direction $d_{(\lightning, a)}$. 
The idea is that direction $d_{(a, e_i)}$ models the effect of applying the $i$th conditional effect of action $a$.
The guard of direction $d_{(a, e_i)}$ ensures that \textbf{(1)} all propositions in $\pre_a$ are true (i.e., the precondition of action $a$ is satisfied), \textbf{(2)} the condition of the conditional effect $e_i$ is satisfied, and \textbf{(3)} variable $\lightning$ is currently set to false (i.e., we have, so far, not played a non-applicable action).
When applying direction $d_{(a, e_i)}$, we change the propositions in $P$ according to $e_i$'s add and delete list (i.e., direction $d_{(a, e_i)}$ sets all variables in $e_i$'s add list to true and all variables in $e_i$'s delete list to false).
Moreover, direction $d_{(a, e_i)}$ sets variable $\mathit{act}_a$ to true and all variables $\mathit{act}_{a'}$ with $a' \neq a$ to false.
The special direction $d_{(\lightning, a)}$ can only be applied if $\pre_a$ does not hold.
If applied, $d_{(\lightning, a)}$ sets variables $\mathit{act}_a$ and $\lightning$ to true. 
As in \Cref{def:enc2}, applying a non-applicable action thus results in a state where $\lightning$ is set to true, prohibiting the future use of directions of the form $d_{(a, e_i)}$.
It is not hard to see that the resulting STS generates all possible executions of $\calP$; for every action $a \in \calO$ and every possible (non-deterministically chosen) conditional effect of $a$, some direction mirrors the same effect in $\calT_\calP$. 
The resulting STS $\calT_\calP$ thus models the same behavior as the explicit-state construction in \Cref{def:enc2}.
As a direct corollary of \Cref{theo:th3}, we get:

\begin{corollary}
	The planning problem $\calP$ admits a conformant plan if and only if $\calT_\calP \models \varphi_\calP$. 
\end{corollary}

Note that we can construct $\calT_\calP$ in linear time in the size of $\calP$. 
It uses $\calO(|P| + |\calO|)$ variables, and the number of directions in $\calT_\calP$ equals the number of total conditional effects in $\calP$.

\paragraph{Implementation}

We have implemented our symbolic encoding in a prototype tool that translates PDDL 2.1 planning problems to HyperLTL verification instances.
Our results show that we translate benchmarks within seconds, indicating that our encoding maintains the core computational challenge.
Current HyperLTL verification tools effectively explore all possible witness paths, leading to poor performance on the resulting instances. 
As we discuss in the next section, we believe that this emphasizes the importance of transferring solutions between both disciplines, and, e.g., study (heuristic) guided approach to HyperLTL model-checking.

\section{Conclusion and Future Work}\label{sec:conc}

In this paper, we have shown that conformant plans and $\exists^*\forall^*$ hyperproperties are closely related.
In particular, conformance is the missing technical link that aligns non-deterministic planning with $\exists^*\forall^*$  model-checking in a sound-and-\emph{complete} (and bi-directional) way.
We hope that this formal connection eventually leads to an improvement in solutions for both problems. 

We empathize that we do not believe that our encodings -- when applied na\"ively -- result in instances that are well-suited for tools in either domain. 
Instead, we view our results as a strong indicator that the fundamental \emph{concepts} developed within either community can be customized to problems in the other community.
For example, the use of heuristics in the verification of hyperproperties is an entirely unexplored, yet very fruitful, direction for future research. 
Current verification approaches to $\exists^*\forall^*$ explore all possible witness paths using, e.g., automata or QBF encodings. 
As our experiments in \Cref{sec:sym2} confirm, such approaches are difficult to scale: even medium-sized conformant planning problems can, when translated into a HyperLTL verification problem, not be solved with current verification tools.
This is unsurprising:
The planning community has demonstrated that exhaustive (non-guided) searches scale poorly but can be improved drastically by employing a (heuristic) guided search.
\Cref{sec:hyper-to-conf} demonstrates that the verification of quantifier alternations in HyperLTL formulas is essentially a planning problem, providing strong evidence that heuristic-guided exploration could lead to much better scalability; even for HyperLTL problems beyond the planning domain.
Using the encoding from \Cref{sec:symbolic}, we could directly translate a HyperLTL verification problem to a planning problem. 
However, we believe that it is much more efficient to directly integrate heuristics into HyperLTL-specific tools by, e.g., guiding the system exploration during bounded model-checking. 

\paragraph*{Acknowledgments}
This work was supported by the European Research Council (ERC) Grant HYPER (101055412), and by the German Research Foundation (DFG) as part of TRR 248 (389792660).

\bibliography{references}

\iffullversion

\newpage

\appendix

\section{Non-Directed Transition Systems}\label{app:ts}

When encoding HyperLTL verification as a conformant planning problem, the actions choosen by the planning agent should determine unique paths for all existentially quantified traces. 
To describe this formally, we used \emph{directions} to unique define transitions within a transition system.
Note that directions are never used within the specification of a system; they merely serve as an auxiliary mechanism to simplify the construction in \Cref{def:enc1}.
For the reverse encoding, we can also work with (standard) non-directed systems:

\begin{definition}\label{def:ts-non-dir}
	A finite-state transition system is a tuple $\calT = (L, l_\mathit{init}, \kappa, \ell)$, where $L$ is a finite set of locations (we use ``locations'' to distinguish them from the ``states'' in a planning domain), $l_\mathit{init} \in L$ is an initial location, $\kappa : L \to 2^L \setminus \{\emptyset\}$ is the transition function, and $\ell : L \to 2^\ap$ labels each location with an evaluation of the APs. 
\end{definition}

Here, the transition function $\kappa$ directly maps to a (non-empty) set of successor states. 
We can easily transform a directed TS (cf.~\Cref{def:ts-dir}) to a TS in \Cref{def:ts-non-dir}, by ignoring all directions. 
Likewise, we can transform a TS from \Cref{def:ts-non-dir} into a directed TS (\Cref{def:ts-dir}) by adding sufficiently many directions; the number of directions is at most $\max_{l \in L} |\kappa(l)|$, i.e., the degree of the TS.

\section{Proofs for \Cref{sec:hyper-to-conf}}\label{app:dir1}

In this section, we prove soundness and completeness of our encoding from \Cref{sec:hyper-to-conf}.

\thsound*

We split \Cref{theo:th1} into two separate theorems (\Cref{theo:e1p1,theo:e1p2}).

\begin{theorem}\label{theo:e1p1}
	If there exists a conformant plan for $\calP_{\calT, \varphi}$, then $\calT \models \varphi$. 
\end{theorem}
\begin{proof}
	Recall that we assume that 
	\begin{align*}
		\varphi = \exists \pi_1\ldots \exists \pi_n. \forall \pi_{n+1}\ldots \forall \pi_{n+m}\ldot \psi
	\end{align*}
	Assume that $\calP_{\calT, \varphi}$ admits a conformant plan and assume that $\langle a_1, \ldots, a_N \rangle$ is such a plan (of length $N$). 
	We will construct witnessing paths $p_1, \ldots, p_n \in \paths(\calT)$ for $\pi_1, \ldots, \pi_n$. 
	Recall that each action in $\calP_{\calT, \varphi}$ is defined by a vector $\vec{d} \in \directions^n$. 
	For each action $a_i$ used in the path and $1 \leq j \leq n$, we write $a_i[j] \in \directions$ for the $j$th direction in vector $a_i$. 
	For each $1 \leq i \leq n$, we now define the \emph{finite} path $r_i \in L^{N+1}$ by
	\begin{align*}
		r_i(0) &:= l_\mathit{init}\\
		r_i(k + 1) &:= \kappa (r_i(k), a_k[i])
	\end{align*}
	That is, we start $r_i$ in the initial location $l_\mathit{init}$ and then apply the directions defined in the the plan (in the $i$th position of the vector).   
	By following the construction, we obtain finite paths $r_1, \ldots, r_n \in L^{N+1}$ (each of the length $N+1$ as the plan defines $N$ actions). 
	As each location has at least one successor location, we can extend $r_1, \ldots, r_n$ to obtain infinite paths, which we call $p_1, \ldots, p_n$.
	We claim that 
	\begin{align*}
		[\pi_1 \mapsto p_1, \ldots, \pi_n \mapsto p_n], 0 \models_\calT  \forall \pi_{n+1}\ldots \forall \pi_{n+m}\ldot \psi,
	\end{align*}
	i.e., the paths are valid concrete choices for the existential quantifiers. 
	
	To show this, assume that $p_{n+1}, \ldots, p_{n+m} \in \paths(\calT)$ are \emph{arbitrary} choices for the universal quantifiers. 
	Define $\Pi := [\pi_1 \mapsto p_1, \ldots, \pi_n \mapsto p_n, \pi_{n+1} \mapsto p_{n+1}, \ldots, \pi_{n+m} \mapsto p_{n+m}]$.
	We claim that $\Pi, 0 \models_\calT \psi$.
	The key observation is that the simulation of $\Pi$ for $N$ steps \emph{is} a path in $\calP_{\calT, \varphi}$ allowed under the fixed plan.
	Formally, we first define the automaton run on $\calA_\psi$ on $\Pi$. 
	Define 
	\begin{align*}
		q_0 &:= q_{0, \psi}\\
		q_{k+1} &:= \delta_\psi(q_k, \bigcup_{i=1}^{n+m} \big\{ (a, \pi_i) \mid a \in \ell(p_i(k)) \})
	\end{align*}
	as the unique run of $\calA_\psi$ on $\Pi$, i.e., in the $k$th step, we read the $k$th states on $p_1, \ldots, p_{n+m}$.
	
	Now consider the following sequences of states in $\calP_{\calT, \varphi}$:
	\begin{align*}
		\langle p_1(0), \ldots, p_{n+m}(0), q_0 \rangle \\
		\langle p_1(1), \ldots, p_{n+m}(1), q_1 \rangle \\
		\cdots \\
		\langle p_1(N), \ldots, p_{n+m}(N), q_N\rangle
	\end{align*}
	We claim that this execution is allowed under plan $\langle a_1, \ldots, a_N \rangle$ in $\calP_{\calT, \varphi}$. 
	Formally, this means that for any $0 \leq k \leq N$, $\langle p_1(k), \ldots, p_{n+m}(k), q_k \rangle \in \llbracket \{s_0\}, \langle a_1, \ldots, a_k \rangle \rrbracket$, i.e., every state in this execution can be reached by executing some prefix plan of $\langle a_1, \ldots, a_N \rangle$.
	The above can be argued inductively:
	Clearly, the first state $\langle p_1(0), \ldots, p_{n+m}(0), q_0 \rangle$ equals $s_0$, the initial state of $\calP_{\calT, \varphi}$. 
	In each step, we have updated $p_1, \ldots, p_n$ using the direction of the plan $\langle a_1, \ldots, a_N \rangle$. 
	At the same time, by definition of $\calP_{\calT, \varphi}$'s effect function, we choose all possible directions to update universally quantified positions. 
	As $p_{n+1}, \ldots, p_{n+m}$ are paths in $\calT$, they can be generated by some sequence of direction, and, as $\calP_{\calT, \varphi}$ explores \emph{all} directions, they will be explored in some non-deterministic outcome. 
	
	So the above execution is allowed under $\langle a_1, \ldots, a_N \rangle$ and we assumed that $\langle a_1, \ldots, a_N \rangle$ is a conformant plan. 
	We thus get that $\langle p_1(N), \ldots, p_{n+m}(N), q_N\rangle \in G$, so, by definition of $\calP_{\calT, \varphi}$, we have $q_N \in F_\psi$. 
	By definition of DFAs, this means that the run of $\calA_\psi$ is accepting. 
	As we assumed that $\calA_\psi$ accepts exactly the words that satisfy $\psi$, so we get that $\Pi, 0 \models_\calT \psi$ as required. 
	As this holds for all possible choices of $p_{n+1}, \ldots, p_{n+m}$, we get that $p_1, \ldots, p_n$ are indeed valid witness paths, so $\calT \models \varphi$, as required. 
\end{proof}

\begin{theorem}\label{theo:e1p2}
	If $\calT \models \varphi$, then there exists a conformant plan for $\calP_{\calT, \varphi}$. 
\end{theorem}
\begin{proof}
	Assume that 
	\begin{align*}
		\varphi = \exists \pi_1\ldots \exists \pi_n. \forall \pi_{n+1}\ldots \forall \pi_{n+m}\ldot \psi
	\end{align*}
	and $\calT \models \varphi$. 
	By the semantics of HyperLTL there thus exists paths $p_1, \ldots, p_n \in \paths(\calT)$ that serve as witnesses for the existential quantifiers, i.e., 
	\begin{align*}
		[\pi_1 \mapsto p_1, \ldots, \pi_n \mapsto p_n], 0 \models_\calT  \forall \pi_{n+1}\ldots \forall \pi_{n+m}\ldot \psi.
	\end{align*}
	Every $p_i$ is generated by at least one infinite sequence of directions. 
	We define $x_1, \ldots, x_n \in \directions^\omega$ as some sequences of directions that generate $p_1, \ldots, p_n$. 
	
	As $p_1, \ldots, p_n$ are witnesses for the existentially quantified paths, for any possible paths $p_{n+1}, \ldots, p_{n+m} \in \paths(\calT)$, we thus get that the combined path assignment $\Pi := [\pi_1 \mapsto p_1, \ldots, \pi_n \mapsto p_n, \pi_{n+1} \mapsto p_{n+1}, \ldots, \pi_{n+m} \mapsto p_{n+m}]$ satisfies $\psi$. 
	By definition of $\calA_\psi$, $\Pi$ is thus accepted by $\calA_\psi$. 
	Formally, for any combination $p_{n+1}, \ldots, p_{n+m}$ of paths, we define $\theta_{p_{n+1}, \ldots, p_{n+m}} \in Q_\psi^\omega$ as the unique run of $\calA_\psi$ on $p_1, \ldots, p_n$ (which are fixed) combined with $p_{n+1}, \ldots, p_{n+m}$.
	As $\theta_{p_{n+1}, \ldots, p_{n+m}}$ is accepting, there thus exists some $n \in \nat$ such that $\theta_{p_{n+1}, \ldots, p_{n+m}}(n) \in F_\psi$. 
	Crucially, as $\calT$ is finite-state and we consider a reachability property, there also exists some global upper bound on a visit of $F_\psi$ (global for all possible choices of $p_{n+1}, \ldots, p_{n+m}$).
	That is, there exists some $N \in \nat$ such that for \emph{every} $p_{n+1}, \ldots, p_{n+m} \in \paths(\calT)$, the run $\theta_{p_{n+1}, \ldots, p_{n+m}}$ visits $F_\psi$ in step $N$ (recall that we assume that accepting states are looping). 
	
	Now define the finite plan $\langle \vec{d}_1, \ldots, \vec{d}_{N-1} \rangle$, where each $\vec{d}_i \in \directions^n$ (the actions in $\calP_{\calT, \varphi}$) is defined by
	\begin{align*}
		\vec{d}_i := (x_1(i), \ldots, x_n(i))
	\end{align*}
	That is, we use the direction sequence that defines $p_1, \ldots, p_n$ as the plan. 
	
	We claim that $\langle \vec{d}_1, \ldots, \vec{d}_{N-1} \rangle$ is a conformant plan. 
	For this, consider any possible execution sequence 
	\begin{align*}
		\langle l_1^0, \ldots, l_{n+m}^{0}, q^0 \rangle \\
		\langle l_1^1, \ldots, l_{n+m}^{1}, q^1 \rangle \\
		\cdots \\
		\langle l_1^{N}, \ldots, l_{n+m}^{N}, q^{N}\rangle
	\end{align*}
	in $\calP_{\calT, \varphi}$ under this plan. 
	By definition of the plan, it is easy to see that $l_i^0 l_i^1 \cdots l_i^{N}$ is a prefix of $p_i$ for each $1 \leq i \leq n$. 
	That is, all existentially quantified transverse exactly the location defined by $p_1, \ldots, p_n$; by design of the plan $\langle \vec{d}_1, \ldots, \vec{d}_N \rangle$. 
	Moreover, by construction of $\calP_{\calT, \varphi}$ is is easy to see that for every $n+1 \leq i \leq n+m$, the location sequence $l_i^0 l_i^1 \cdots l_i^{N}$ is a prefix of some path in $\calT$ (as the effect function of $\calP_{\calT, \varphi}$ uses $\calT$'s transition function). 
	Let $\dot{p}_{n+1}, \ldots, \dot{p}_{n+m} \in \paths(\calT)$ be any such paths, i.e., for every $n+1 \leq i \leq n+m$, $l_i^0 l_i^1 \cdots l_i^{N}$ is a prefix of $\dot{p}_i$. 
	Now by the construction of the concrete witnesses $p_1, \ldots, p_n$, we get that $\theta_{\dot{p}_{n+1}, \ldots, \dot{p}_{n+m}}$ is an accepting run of $\calA_\psi$. 
	That is, $p_1, \ldots, p_n$ together with $\dot{p}_{n+1}, \ldots, \dot{p}_{n+m}$ satisfies $\psi$. 
	The sequence of automaton states $q^0q^1\cdots q^{N}$ is -- by design of $\calP_{\calT, \varphi}$ -- exactly a prefix of $\theta_{\dot{p}_{n+1}, \ldots, \dot{p}_{n+m}}$ (in $\calP_{\calT, \varphi}$'s effect function, we always update the automaton state based on $\calA_\psi$'s deterministic transition function). 
	As $\theta_{\dot{p}_{n+1}, \ldots, \dot{p}_{n+m}}(N) \in F_\psi$ (by the choice of $N$), we get that $q^N \in F_\psi$.
	By construction of $\calP_{\calT, \varphi}$, this implies that $\langle l_1^{N}, \ldots, l_{n+m}^{N}, q^{N}\rangle$ (the last state in the execution sequence of the plan), is a goal state of $\calP_{\calT, \varphi}$.
	As this holds for all possible executions of $\langle \vec{d}_1, \ldots, \vec{d}_N \rangle$, we get that $\langle \vec{d}_1, \ldots, \vec{d}_N \rangle$ is indeed a conformant plan for $\calP_{\calT, \varphi}$ as required.
\end{proof}

\section{Proofs for \Cref{sec:conf-to-hyper}}\label{app:dir2}

In this section, we prove soundness and completeness of our encoding from \Cref{sec:conf-to-hyper}.

\second*

We split the proof of \Cref{theo:th3} into two separate propositions (\Cref{prop:p1,prop:p2}); one for each implication:

\begin{theorem}\label{prop:p1}
	If $\calP$ admits a conformant plan, the $\calT_\calP \models \varphi_\calP$.
\end{theorem}
\begin{proof}
	Assume that $\calP = (S, s_0, G, \calO)$ and recall that $\varphi_\calP$ is defined as
	\begin{align*}
		\exists \pi_1\ldot \forall \pi_2\ldot \ltlF \mathit{goal}_{\pi_2} \lor  \ltlF \Big(\bigvee_{a \in \calO} (\mathit{act}_a)_{\pi_1} \not\leftrightarrow (\mathit{act}_a)_{\pi_2} \Big).
	\end{align*}
	Now assume that $\langle a_1, \ldots, a_n \rangle$ is a conformant plan in $\calP$. 
	Further assume that $s_0s_1 \cdots s_n \in S^*$ is some execution in $\calP$ under this plan, i.e.,  $s_0$ is $\calP$'s initial state, and for each $0 < i < n$, $s_{i-1} \in \pre_{a_{i}}$ and $s_{i} \in \eff_{a_{i}}(s_{i-1})$. 
	
	Consider the finite sequence 
	\begin{align*}
		(s_0, a_0)(s_1, a_1) \cdots (s_n, a_n) \in (S \times \calO)^*,
	\end{align*}
	where $a_0 \in \calO$ is an arbitrary action. 
	It is easy to see that -- by definition of $\calT_\calP$ -- this sequence is the prefix of some infinite path in $\calT_\calP$: 
	In $\calT_\calP$, the transition function exactly allows the transition permitted by $\calP$'s action effects. 
	Let ${p} \in \paths(\calT_\calP)$ be any such path that has this sequence of state-action pairs as a prefix. 
	To prove $\calT_\calP \models \varphi_\calP$, we claim that 
	\begin{align*}
		[\pi_1 \mapsto {p}], 0 \models_{\calT_\calP}  &\forall \pi_2\ldot \ltlF \mathit{goal}_{\pi_2} \lor  \\
		&\quad \ltlF \Big(\bigvee_{a \in \calO} (\mathit{act}_a)_{\pi_1} \not\leftrightarrow (\mathit{act}_a)_{\pi_2} \Big),
	\end{align*} 
	i.e., ${p}$ is a valid witness for the existentially quantified $\pi_1$. 
	To show this, assume that $p' \in \paths(\calT_\calP)$ is an arbitrary path (for $\pi_2$).
	We claim that $[\pi_1 \mapsto {p}, \pi_2 \mapsto p']$ satisfy the LTL body of $\varphi_\calP$.
	For this, we can assume that $[\pi_1 \mapsto {p}, \pi_2 \mapsto p']$ satisfies
	\begin{align}\label{eq:assumption}
		\ltlG \Big(\bigwedge_{a \in \calO} (\mathit{act}_a)_{\pi_1} \leftrightarrow (\mathit{act}_a)_{\pi_2} \Big),
	\end{align}
	i.e., both paths always agree on the $\mathit{act}_a$ propositions: If this is not the case, the second disjunct in $\varphi_\calP$'s body is satisfied so the formula holds trivially.
	
	We now look at the first $n+1$ steps on $p'$, yielding a finite sequence 
	\begin{align*}
		(s_0', a_0')(s_1', a_1') \cdots (s_n', a_n')
	\end{align*}
	of state-action pairs.
	Here, each $a_i' \in \calO$, and each $s_i'$ is either a state in $\calP$ or equals the error location $\lightning$ (cf.~\Cref{def:enc2}).
	By the design of $\calT_\calP$'s labeling function (which always sets the $\mathit{act}_a$ APs based on the action in each state-action pair), we get that the action sequence in $p'$ is the same as the action sequence $p$, i.e., $\langle a_0', a_1', \ldots, a_n' \rangle =  \langle a_0, a_1, \ldots, a_n \rangle$.
	By $\calT_\calP$'s transition function, this implies that $s_0's_1'\cdots s_n'$ is some execution of $\calP$ under plan $\langle a_0', a_1', \ldots, a_n' \rangle =  \langle a_0, a_1, \ldots, a_n \rangle$. 
	Firstly, this already ensures that $s_i' \neq \lightning$ (as in a conformant plan, we never encounter a situation where an action is not applicable). 
	Moreover (again, as $\langle a_0, a_1, \ldots, a_n \rangle$ is, by assumption, a conformant plan), we get that $s_n' \in G$, i.e., the exeuction ends in a goal state. 
	By $\calT_\calP$'s labeling function, we thus get that $\mathit{goal} \in \ell\big((s_n', a_n')\big)$; path $p'$ thus visits a state labeling with AP $\mathit{goal}$.
	This implies that $[\pi_1 \mapsto p, \pi_2 \mapsto p']$ satsifies $\ltlF \mathit{goal}_{\pi_2}$. 
	Consequently, for \emph{every} $p' \in \paths(\calT_\calP)$, the path assignment $[\pi_1 \mapsto p, \pi_2 \mapsto p']$ satisfies $\varphi_\calP$'s body (either by violating \ref{eq:assumption} or satisfying $\ltlF \mathit{goal}_{\pi_2}$).
	So $\calT_\calP \models \varphi_\calP$ as required.
\end{proof}

\begin{table}[!t]
	\caption{We translate the conformant planning benchmarks from \citet{HoffmannB06} to HyperLTL benchmarks using our prototype tool.
		We report the runtime in seconds (of the translation) and the number of Boolean variables used in the NuSMV model. }\label{tab:exp}
	\centering
	\begin{tabular}{lll}
		\toprule
		\textbf{Name} & \textbf{$|$Vars$|$} & \textbf{t} \\
		\midrule  
		\textsc{Block-2-5} & 45 & 0.1 \\
		\textsc{Block-2-6} & 59 & 0.1 \\
		\textsc{Block-2-7} & 74 & 0.2 \\
		\textsc{Block-2-13} & 209 & 1.8 \\
		\textsc{Block-2-20} & 456 & 19.2 \\
		\textsc{Block-3-5} & 45 & 0.2 \\
		\textsc{Block-3-6} & 59 & 0.2 \\
		\textsc{Block-3-7} & 74 & 0.3 \\
		\textsc{Block-3-13} & 209 & 1.8 \\
		\textsc{Block-3-20} & 456 & 19.4 \\
		\textsc{Block-4-5} & 45 & 0.2 \\
		\textsc{Block-4-6} & 59 & 0.1 \\
		\textsc{Block-4-7} & 74 & 0.3 \\
		\textsc{Block-4-13} & 209 & 1.9 \\
		\textsc{Block-4-20} & 456 & 19.1 \\
		\midrule
		\textsc{Bomb-5-1} & 11 & 0.1 \\
		\textsc{Bomb-5-5} & 17 & 0.1 \\
		\textsc{Bomb-5-10} & 23 & 0.1 \\
		\textsc{Bomb-10-1} & 17 & 0.1 \\
		\textsc{Bomb-10-5} & 23 & 0.1 \\
		\textsc{Bomb-10-10} & 29 & 0.1 \\
		\textsc{Bomb-20-1} & 28 & 0.1 \\
		\textsc{Bomb-20-5} & 34 & 0.1 \\
		\textsc{Bomb-20-10} & 40 & 0.1 \\
		\textsc{Bomb-20-20} & 51 & 0.2 \\
		\textsc{Bomb-50-1} & 59 & 0.1 \\
		\textsc{Bomb-50-5} & 65 & 0.1 \\
		\textsc{Bomb-50-10} & 71 & 0.1 \\
		\textsc{Bomb-50-50} & 114 & 1.2 \\
		\textsc{Bomb-100-1} & 110 & 0.1 \\
		\textsc{Bomb-100-5} & 116 & 0.1 \\
		\textsc{Bomb-100-10} & 122 & 0.4 \\
		\textsc{Bomb-100-60} & 175 & 4.2 \\
		\textsc{Bomb-100-100} & 216 & 9.7 \\
		\midrule
		\textsc{Logistics-2-2-2} & 35 & 0.1 \\
		\textsc{Logistics-2-2-4} & 49 & 0.1 \\
		\textsc{Logistics-2-3-2} & 54 & 0.4 \\
		\textsc{Logistics-2-3-3} & 73 & 0.4 \\
		\textsc{Logistics-3-2-2} & 54 & 0.2 \\
		\textsc{Logistics-3-2-4} & 74 & 0.2 \\
		\textsc{Logistics-3-3-3} & 98 & 1.2 \\
		\textsc{Logistics-4-2-2} & 56 & 0.4 \\
		\textsc{Logistics-4-2-4} & 79 & 0.5 \\
		\textsc{Logistics-4-3-2} & 91 & 3.5 \\
		\textsc{Logistics-4-3-3} & 123 & 3.6 \\
		\bottomrule
	\end{tabular}

\end{table}

\begin{theorem}\label{prop:p2}
	If $\calT_\calP \models \varphi_\calP$, then $\calP$ admits a conformant plan.
\end{theorem}
\begin{proof}
	Assume that $\calT_\calP \models \varphi_\calP$.
	So we can find at least one $p \in \paths(\calT_\calP)$ as a witness for the existentially quantified path $\pi_1$, i.e.,
	\begin{align*}
		[\pi_1 \mapsto {p}], 0 \models_{\calT_\calP}  &\forall \pi_2\ldot \ltlF \mathit{goal}_{\pi_2} \lor  \\
		&\quad \ltlF \Big(\bigvee_{a \in \calO} (\mathit{act}_a)_{\pi_1} \not\leftrightarrow (\mathit{act}_a)_{\pi_2} \Big).
	\end{align*} 
	Our first observation is that the length up to which the goal state must be reached is bounded by some constants. 
	That is, define the set 
	\begin{align*}
		T := \Big\{  p' \in \paths(\calT_\calP) \mid &[\pi_1 \mapsto p, \pi_2 \mapsto p'], 0 \models \\
		&\ltlG \big(\bigwedge_{a \in \calO} (\mathit{act}_a)_{\pi_1} \leftrightarrow (\mathit{act}_a)_{\pi_2} \big) \Big\}
	\end{align*}
	as the set of all paths in $\calT_\calP$ that have the same sequence of actions as $p$ (i.e., violate the second disjunct in $\varphi_\calP$). 
	As $p$ is a valid witness for the existential quantifier, we thus get that for every $p' \in T$, we have $[\pi_1 \mapsto p, \pi_2 \mapsto p'], 0 \models  \ltlF \mathit{goal}_{\pi_2}$. 
	As $\calT_\calP$ contains only finitely many states, there must exists a global bound on the first visit of a state labeled with AP $\mathit{goal}$. 
	That is, there exists some $N \in \nat$, such that for every $p' \in T$, we have $\mathit{goal} \in \ell(p'(N-1))$ (If such a bound would not exists, we could find a lass-shaped trace in $\calT_\calP$ that never visits the goal). 
	
	Now the fixed witnes paths $p$ has the form $p = (s_0, a_0) (s_1, a_1) \cdots$. 
	We define the plan $\langle a_1, a_2, \ldots, a_{N-1} \rangle$ by using the first $N$ actions used in $p$. 
	We claim that is plan is a conformant plan in $\calP$. 
	To show this, we consider any sequence $s_0's_1'\ldots s'_{N-1}$ of states in $\calP$ under this plan. 
	The key is that, by definition of $\calT_\calP$, there exists some path in $\paths(\calT_\calP)$ that has $(s_0', a_0) (s_1', a_1) \cdots (s'_{N-1}, a_{N-1})$ as a prefix, i.e., uses the same sequence of actions as in the plan and traverses the same states as in $s_0's_1'\ldots s'_{N-1}$. 
	Moreover, there also exists some path $p'$ that has this sequence as a prefix and is contained in $T$ (after the first $N$ steps, we still use the same sequence of actions as in $p$). 
	Firstly, this implies that, during the execution of $\langle a_1, a_2, \ldots, a_{N-1} \rangle$, we never apply an action that is not-applicable (as this would, in $\calT_\calP$ result in an error location which would violate $\varphi_\calP$ as no location with AP $\mathit{goal}$ can be reached from an error location). 
	As the path $p'$ is contained in $T$, it visits a state labeled with AP $\mathit{goal}$ in the last step (i.e, the $N$the step). 
	By definition of $\calT_\calP$'s labeling function, this implies that the last state $s_{N-1}$ is contained in $\calP$'s goal states. 
	
	Consequently, we have prove that every execution of $\calP$ under plan $\langle a_1, a_2, \ldots, a_{N-1} \rangle$ ends in a goal state. 
	The plan $\langle a_1, a_2, \ldots, a_{N-1} \rangle$  is thus a conformant plan, as required.
\end{proof}

\section{Prototype Implementation}\label{app:imp}

In our paper, we always work with \emph{explicitly} represented transition system and \emph{explicit-state} represented planning problems. 
In practice, planning tools work with a symbolically represented system (e.g., in STRIPS or PDDL) and, likewise, transition systems are represented using circuits or Boolean variables. 

To demonstrate that our approach also applies to symbols systems, we have implemented the encoding from \Cref{sec:conf-to-hyper} into a prototype.
Our prototype takes a PDDL 2.1 planning problem (with non-deterministic action effects and non-deterministic initial states) and translates it to a NuSMV system \cite{CimattiCGGPRST02} and HyperLTL formula. 
Our tool uses a custom grounder with a few optimizations and converts the grounded instance into a Boolean transition system, which we then export in the NuSMV format.
We use the formats used by HyperLTL verification tools such as \texttt{HyperQB} \cite{HsuSB21}, \texttt{McHyper} \cite{FinkbeinerRS15}, and \texttt{AutoHyper} \cite{BeutnerF23}. 

We use the conformant planning benchmarks shipped with \texttt{ConformantFF} \cite{HoffmannB06} and translate each instance using our tool. 
We report the runtime and the number of Boolean variables on a selected of 3 popular domains in \Cref{tab:exp} (running on an Macbook with M1 Pro CPU).
We observe that our tool can translate all benchmarks within seconds. 
This also provides evidence that our encoding really encodes the ``computational essence'' of conformant planning into HyperLTL verification. 
In particular, it does not ``pre-solve'' parts of the problem, but really yields an (at least) equally changeling model-checking problem.

\fi

\end{document}